\documentclass[letterpaper, 10 pt, conference]{ieeeconf}  

\usepackage{graphicx}
\usepackage{subcaption} 
\usepackage{caption}
\usepackage{glossaries}

\usepackage{amssymb}
\usepackage{amsmath}
\usepackage{commath}
\usepackage{mathtools}

\usepackage{hyperref}
\hypersetup{
	colorlinks=true,
	linkcolor=[rgb]{0.015, 0.59, 0.77},
	citecolor=[rgb]{0.78, 0.03, 0},  
	urlcolor=[rgb]{1, 0.612, 0.03}
}
\usepackage{import}
\usepackage{paralist}
\usepackage{gensymb}
\usepackage{dsfont}
\usepackage[separate-uncertainty = true,multi-part-units=single]{siunitx}
\usepackage[bottom]{footmisc}
\usepackage{framed}
\usepackage{balance} 
\usepackage{color, colortbl}
\usepackage[dvipsnames]{xcolor}
\usepackage[skins]{tcolorbox}
\usepackage{xurl}
\usepackage{microtype}
\usepackage{tikzscale} 
\usepackage{mathrsfs}

\usepackage{nameref}

\makeglossaries
\newacronym{MANN}{MANN}{Mode-Adaptive Neural Networks}
\newacronym{mocap}{MoCap}{motion capture}
\newacronym{com}{CoM}{center of mass}
\newacronym{qp}{QP}{Quadratic Programming}
\newacronym{mpc}{MPC}{Model Predictive Control}
\newacronym{rhp}{RHP}{Receding Horizon Planner}
\newacronym{zmp}{ZMP}{zero moment point}
\newacronym{ik}{IK}{inverse kinematics}
\newacronym{dnn}{DNN}{Deep Neural Network}
\newacronym{imu}{IMU}{inertial measurement unit}
\newacronym{cbf}{CBF}{control barrier function}
\newacronym{ga}{GA}{genetic algorithm}
\newacronym{ddp}{DDP}{differential dynamic programming}
\newacronym{kf}{KF}{Kalman filter}

\DeclareMathOperator*{\argmax}{arg\,max}

\usepackage{eso-pic}
\usepackage{graphicx} 

\AddToShipoutPictureBG*{%
  \AtPageUpperLeft{%
    \setlength{\unitlength}{1mm}%
    \put(0,-12){\makebox(\paperwidth,0)[c]{\parbox{0.8\textwidth}{\centering\textcolor{gray}{\large This paper has been accepted for publication at 2024 IEEE-RAS International Conference on Humanoid Robots, ©IEEE}}}}
  }
}

\IEEEoverridecommandlockouts                              

\title{\LARGE \bf
Online DNN-driven Nonlinear MPC\\for Stylistic Humanoid Robot Walking with Step Adjustment
}

\author{Giulio Romualdi$^{\dagger,1}$, Paolo Maria Viceconte$^{\dagger,1}$, Lorenzo Moretti$^{1}$, Ines Sorrentino$^{1,2}$, \\
Stefano Dafarra$^{1}$, Silvio Traversaro$^{1}$ and Daniele Pucci$^{1,2}$
\thanks{$^{\dagger}$ Paolo Maria Viceconte and Giulio Romualdi are co-first authors.}
\thanks{$^{1}$ Artificial and Mechanical Intelligence, Istituto Italiano di Tecnologia, Genoa, Italy {\tt\small name.surname@iit.it}}
\thanks{$^{2}$ School of Computer Science, University of Manchester, Manchester, United Kingdom}
\thanks{This paper was supported by the Italian National Institute for Insurance against Accidents at Work (INAIL) ergoCub Project.}}

\begin{document}

\bstctlcite{IEEEexample:BSTcontrol}

\maketitle
\thispagestyle{empty}
\pagestyle{empty}

\begin{abstract}

This paper presents a three-layered architecture that enables stylistic locomotion with online contact location adjustment. Our method combines an autoregressive Deep Neural Network (DNN) acting as a \emph{trajectory generation} layer with a model-based \emph{trajectory adjustment} and \emph{trajectory control} layers.
The DNN produces centroidal and postural references serving as an initial guess and regularizer for the other layers. 
Being the DNN trained on human motion capture data, the resulting robot motion exhibits locomotion patterns, resembling a human walking style.
The trajectory adjustment layer utilizes non-linear optimization to ensure dynamically feasible center of mass (CoM) motion while addressing step adjustments. We compare two implementations of the trajectory adjustment layer: one as a receding horizon planner (RHP) and the other as a model predictive controller (MPC). To enhance MPC performance,
we introduce a Kalman filter to reduce measurement noise. The filter parameters are automatically tuned with a Genetic Algorithm.
Experimental results on the ergoCub humanoid robot demonstrate the system's ability to prevent falls, replicate human walking styles, and withstand disturbances up to 68 Newton.\looseness=-1

\end{abstract}

\section{Introduction}

Designing trajectories for humanoid robots involves addressing complex interactions between high-dimensional multi-body systems and the environment. Additionally, in scenarios where unexpected disturbances affect the robot, it becomes essential to adapt the planned contact position to prevent falls. 
There are two fundamental approaches to control humanoid robot locomotion: \emph{data-driven} and \emph{model-based}. 
The data-driven approach involves designing a reinforcement learning infrastructure that takes high-level commands and the robot's state as input, generating joint references. This architecture produces natural robot motions~\cite{Radosavovic2023Real-WorldLearning,Li2024ReinforcementControl}, 
but it suffers from the sim-to-real gap~\cite{Kim2023Torque-BasedTransfer}. On the other hand, the model-based approach usually involves a hierarchical architecture with three main layers~\cite{Romualdi2020ARobots}. Each layer generates references for the inner layers by processing inputs from the robot, the environment, and the outer layers. 
From top to bottom, these layers are referred to as \emph{trajectory generation}, \emph{trajectory adjustment}, and \emph{trajectory control}.
\par
This paper aims to bridge the gap between the data-driven and the model-based approaches, proposing a hierarchical control architecture where a \gls{dnn} acts as a trajectory generation layer and a \gls{mpc} serves as a trajectory adjustment layer. This infrastructure enables stylistic human-like locomotion with online step adjustment capabilities.
\begin{figure}[t]
\vspace{0.2cm}
  \centering
      \begin{subfigure}[b]{0.31\columnwidth}
        \centering
        \includegraphics[width=\textwidth]{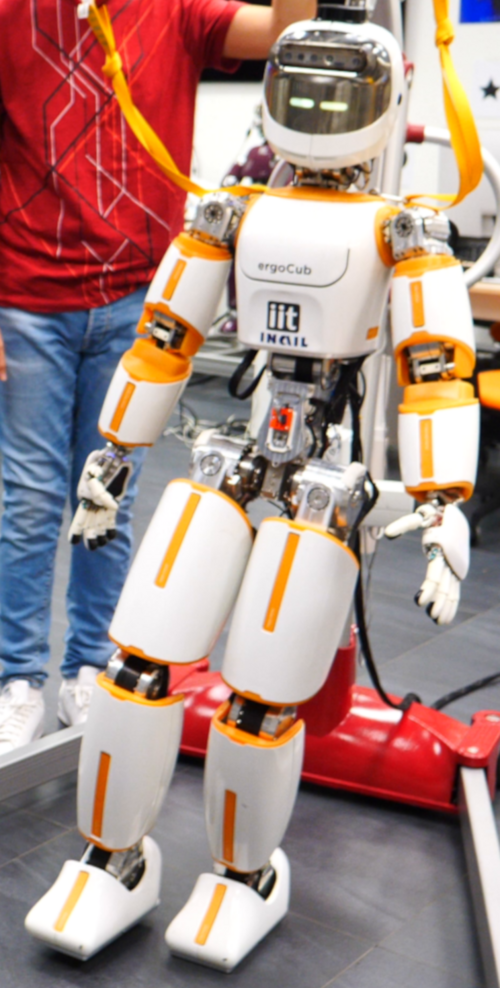}
    \end{subfigure}
    \hfill
     \begin{subfigure}[b]{0.31\columnwidth}
        \centering
        \includegraphics[width=\textwidth]{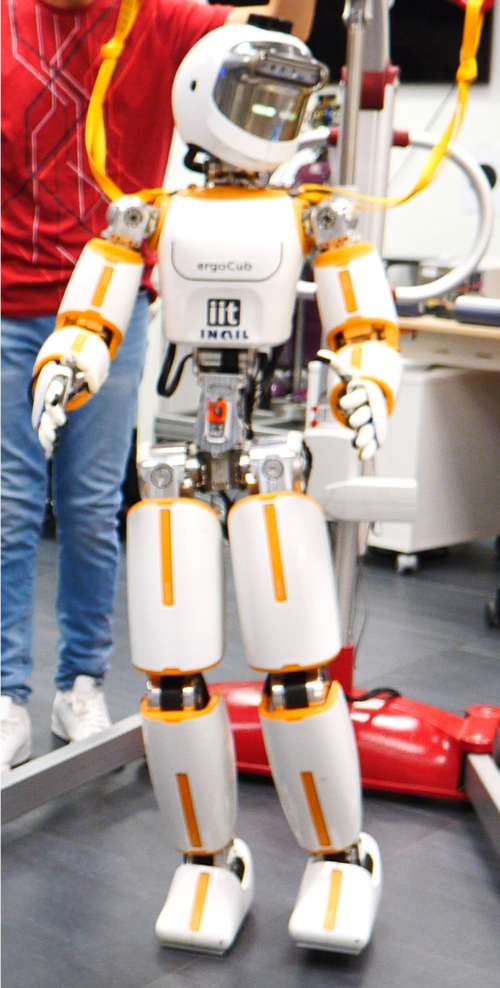}
    \end{subfigure}
    \hfill
    \begin{subfigure}[b]{0.31\columnwidth}
        \centering
        \includegraphics[width=\textwidth]{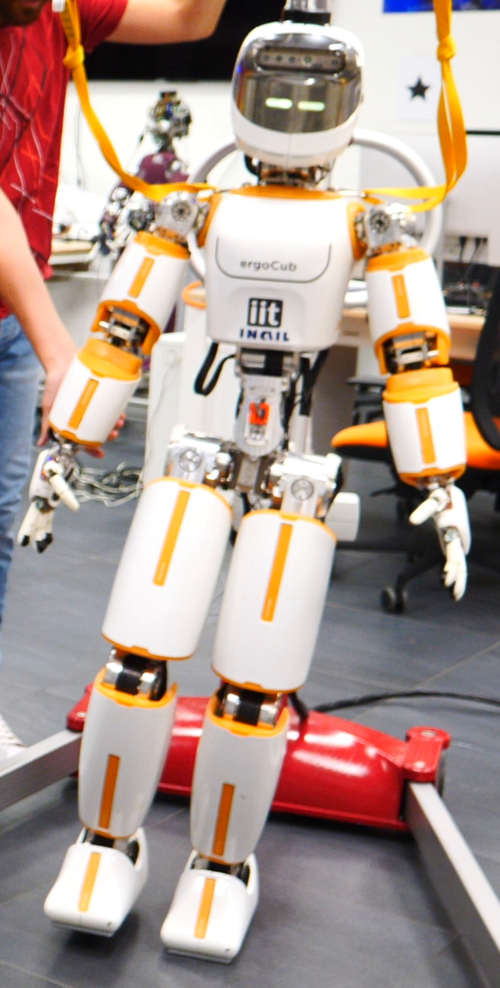}
    \end{subfigure}
  \caption{The humanoid robot ergoCub walks forward, sideways, and reacts to an external force with the proposed locomotion architecture.}  
    \vspace{-0.5cm}
    \label{fig:ergoCub_walks}
\end{figure}

Within the model-based architecture, the \emph{trajectory generation} layer is responsible for computing a sequence of contact locations and timings, along with joint and \gls{com} trajectories.
This layer can be designed using \emph{model-based} or \emph{learning-based} methods. In the model-based approach, optimization techniques are applied to evaluate the feasibility of contact locations, utilizing both whole-body robot models~\cite{dai2014whole} or template models~\cite{Takenaka2009RealGeneration-,Englsberger2015}. While providing enhanced planning capabilities, a whole-body trajectory generator may lead to extended computation time~\cite{Nguyen2020DynamicFunctions}, preventing their use in real-time. Predefined contact locations and timings allow whole-body planners to operate in real-time~\cite{Dantec2022FirstFeedback}; however, in case of external disturbances, the contact position is not adapted.
Instead, limiting the robot \gls{com} height and the angular momentum during walking simplifies the robot model into template models, making the problem treatable online~\cite{Takenaka2009RealGeneration-,Englsberger2019}. However, 
these architectures tend to neglect stylistic properties of the generated motions, as they are challenging to explicitly encode and optimize for~\cite{Hsu2005StyleMotion,Du2019StylisticModels}. 
On the other hand, learning-based approaches leverage data-driven techniques to transfer stylistic locomotion patterns to the robot. These methods often utilize human \gls{mocap} data to train \gls{dnn}s~\cite{bergamin_drecon_2019,starke_deepphase_2022}.\looseness=-1

The \emph{trajectory adjustment} layer considers the output of the trajectory generation layer and the robot state to adjust contact locations and the \gls{com} trajectory. 
This layer often relies on template models that consider a constant~\cite{Stephens2010PushJoints,Shafiee2019OnlineRobots}
or adaptive~\cite{Griffin2017WalkingAtlas,Scianca2020MPCFeasibility} step duration. However, simplified model-based controller architectures often treat footstep adjustment separately from the primary control loop~\cite{Kim2023FootControl,Dallard2024RobustStabilizers}.
An alternative approach involves framing the problem as a whole-body \gls{mpc} and solving it using \gls{ddp}~\cite{Shim2023Topology-BasedSelection}. Given the complexity of the problem, it is necessary to correctly warm-start the \gls{ddp} algorithm to obtain reliable results~\cite{Dantec2021WholeTalos}.
Conversely, one can model the robot as a single rigid body~\cite{Grandia2021Multi-LayeredControl} or consider its centroidal dynamics~\cite{Orin2013} to deal with the step adjustment within the main control loop~\cite{Ding2022Orientation-AwareWalking,Meng2024OnlineExtension}.

Finally, the \emph{trajectory control} layer generates robot commands to stabilize the references produced by its preceding layers. This layer considers whole-body kinematics or dynamics. The associated optimization problem is typically structured with either strict or weighted stack of task~\cite{Stephens2010PushJoints,Caron2019StairControl} and framed as a \gls{qp} problem.

This paper presents a solution for online whole-body trajectory generation within a model-based control framework by leveraging data-driven methods. Specifically, the \gls{dnn} extracts references from \gls{mocap} data to initialize the trajectory-adjustment layer. 
This architecture enables stylistic-human-like locomotion patterns with online step adjustment capabilities.
In detail, our contributions are:
i) An architecture that connects a \gls{dnn} trajectory generator with a \gls{mpc}-based trajectory adjustment layer. This approach is demonstrated using \gls{MANN}~\cite{Viceconte2022ADHERENT:Robots} as \gls{dnn}.
ii) A \gls{mpc} formulation ensuring kinematically-feasible motions of the robot's \gls{com} while addressing step adjustment. Unlike previous approaches~\cite{Meng2024OnlineExtension,Romualdi2022OnlineAdjustment}, our proposed controller uses a control-barrier function to ensure that the \gls{com} remains within a safe region.
iii) Benchmarking of two implementations of trajectory adjustment layer acting as a \gls{rhp} and as a \gls{mpc}, respectively. In the latter case, we introduce a \gls{kf}~\cite{Kalman1960AProblems} to reduce the noise of \gls{com} velocity and angular momentum measurements. \gls{kf} parameters are optimized via a \gls{ga}.
The system is validated on the position-controlled humanoid robot ergoCub -- Fig.~\ref{fig:ergoCub_walks}. 
We demonstrate the effectiveness of the proposed architecture in preventing the robot from falling while walking, when subject to impulsive disturbances up to \SI{68}{\newton}. The style of the robot motion echoes the \gls{mocap} data.
\par
The paper is organized as follows. Sec.\ref{sec:background} introduces notation and background. Sec.\ref{sec:method} presents the layers of the proposed architecture and their interconnection. Sec.\ref{sec:results} shows the results and their discussion. Sec.\ref{sec:conclusions} concludes the paper.

\section{Background}
\label{sec:background}
\subsection{Notation}
\begin{itemize}
\item $I_n$ and $0_n$ represent $n \times n$ identity and zero matrices; 
\item $e_i$ indicates the canonical base, i.e., $e_3^\top = \begin{bmatrix}0& 0& 1\end{bmatrix}$
\item $\mathcal{I}$ denotes the inertial frame;
\item $\prescript{\mathcal{A}}{}{p}_\mathcal{C}$ is a vector that connects the origin of frame $\mathcal{A}$ and the origin of frame $\mathcal{C}$ expressed in the frame $\mathcal{A}$;
\item Given $\prescript{\mathcal{A}}{}{p}_\mathcal{C}$ and $\prescript{\mathcal{B}}{}{p}_\mathcal{C}$,  $\prescript{\mathcal{A}}{}{p}_\mathcal{C} = \prescript{\mathcal{A}}{}{R}_\mathcal{B} \prescript{\mathcal{B}}{}{p}_\mathcal{C} + \prescript{\mathcal{A}}{}{p}_\mathcal{B} = \prescript{\mathcal{A}}{}{H}_\mathcal{B}\prescript{\mathcal{B}}{}{p}_\mathcal{C}$, where $\prescript{\mathcal{A}}{}{H}_\mathcal{B}$ is the homogeneous transformation and $\prescript{\mathcal{A}}{}{R}_\mathcal{B} \in SO(3)$ is the rotation matrix; 
\item the \emph{hat operator} is $^\wedge : \mathbb{R}^3  \to \mathfrak{so}(3)$,  such that
$x^\wedge y = x \times y$. $\times$ is the cross product operator in $\mathbb{R}^3$;
\item $\prescript{\mathcal{A}}{}{{v}}_\mathcal{B} \in \mathbb{R}^3$ is the time derivative of the relative position between the origin of the frame $\mathcal{B}$ and $\mathcal{A}$, $\prescript{\mathcal{A}}{}{{v}}_\mathcal{B}  = \prescript{\mathcal{A}}{}{\dot{p}}_\mathcal{B}$;
\item $\prescript{\mathcal{A}}{}{{f}}_\mathcal{B} \in \mathbb{R}^3$ is the force acting on $\mathcal{B}$ expressed in $\mathcal{A}$;
\item for the sake of clarity, the prescript $\mathcal{I}$ will be omitted.
\end{itemize}

\subsection{Centroidal dynamics}

\begin{figure*}[t]
\resizebox{\textwidth}{!}{%
    \centering\includegraphics{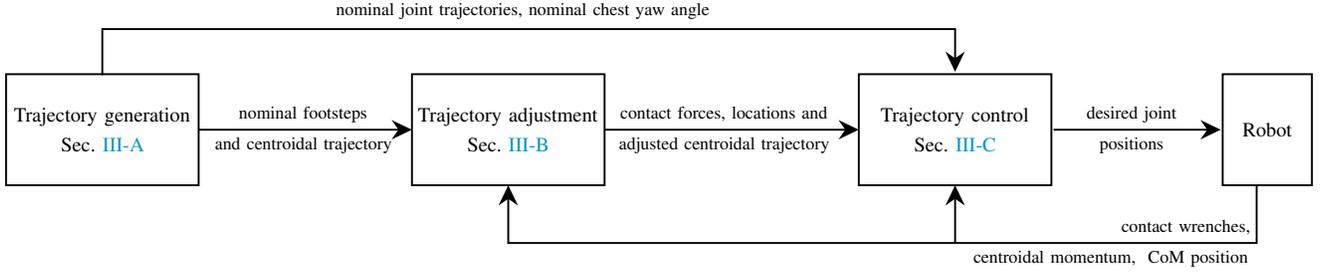} 
    }
    \caption{The proposed three-layer hierarchical architecture integrating a \gls{dnn} as a trajectory generation layer and a model-based trajectory adjustment layer.}
    \label{fig:architecture}
    \vspace{-0.3cm}
\end{figure*}

The centroidal momentum ${}_{\bar{G}}h ^\top= \begin{bmatrix}
{}_{\bar{G}}h^{p\top} & {}_{\bar{G}}h^{\omega\top}
\end{bmatrix} \in \mathbb{R}^6$ is the combined linear and angular momentum of each robot link, relative to the robot's \gls{com}~\cite{Orin2013}. The vectors ${}_{\bar{G}} h^p \in \mathbb{R}^3$ and ${}_{\bar{G}} h^\omega \in \mathbb{R}^3$ denote the linear and angular momentum, respectively. The coordinates of ${}_{\bar{G}} h$ are expressed with respect to a frame centered at the robot's \gls{com} and oriented as $\mathcal{I}$~\cite{Orin2013}. 
In the case of rectangular contact surfaces the time derivative of the centroidal momentum writes as
\begin{equation}
{}_{\bar{G}}\dot{h} = \sum_{i = 1}^{n_c} \sum_{j = 1}^{4} \begin{bmatrix}
I_3 \\
({H}_{p_{C_i}} p_{v_{i,j}} -p_{\text{CoM}})^\wedge
\end{bmatrix} f_{i,j} + \begin{bmatrix} m g \\ 0_{3\times1} \end{bmatrix}.
\label{eq:centroidal_dynamics}
\end{equation}
Where $m$ is the robot's mass, $n_c$ is the number of active contacts.
$p_{v_{i,j}}$ is the position of vertex $j$ of contact $i$, expressed with respect to the frame associated with the contact surface. ${H}_{p_{C_i}}$ is the pose of the contact $i$. $f_{i,j}\in\mathbb{R}^3$ denotes the pure force applied to vertex $j$ of contact $i$.

\subsection{Control Barrier Function}
Let us consider a discrete dynamical system $x_{k+1} = f(x_k, u_k)$ where $x_k \in X \subseteq \mathbb{R}^n $ is the system state and $u_k\in U$ is the control input. We define a \emph{safe set} $\Omega = \{ x\in X : g(x) \geq 0 \}$ as the superlevel set of a continuously-differentiable function $g : X \subseteq \mathbb{R}^n \rightarrow \mathbb{R}$. $g$ is a discrete-time \gls{cbf}~\cite{Sun2023Safety-CriticalControl}  if satisfies
\begin{equation}
    \Delta g (x_{k+1}, x_k)  \geq -\gamma g(x_k),
    \label{eq:cbf_constraint}
\end{equation}
where $\Delta g (x_{k+1}, x_k) = g(x_{k+1}) - g(x_k)$, with $0 < \gamma \leq 1$.

\subsection{Problem statement}
\label{sec:problem_statement}



Let us consider the architecture in Fig.~\ref{fig:architecture}, where an autoregressive \gls{dnn} generates trajectories and a \gls{mpc} adjusts them dynamically.
The goal is to seamlessly integrate these layers, allowing the \gls{mpc} to use the \gls{dnn}-generated trajectories as a warm start, despite their potential lack of dynamic feasibility~\cite{Viceconte2022ADHERENT:Robots}. This integration faces several challenges:

\begin{itemize}
    \item \label{c1} $\mathscr{C}_1$: the \gls{dnn}-based trajectory generation lacks contact awareness, whereas the \gls{mpc} needs contact locations and timings. 
    \item \label{c2} $\mathscr{C}_2$: the \gls{dnn} provides next-instant outputs, but the \gls{mpc} needs a trajectory over a time horizon. 
    \item \label{c3} $\mathscr{C}_3$: the \gls{dnn} is trained with a dataset collected at a specific frequency, while the \gls{mpc} may expect references at a different rate.
\end{itemize}

These challenges are addressed by designing ad-hoc blocks to extract contact information, generate trajectories for the time horizon, and match reference frequencies, respectively, as detailed in Sec.~\ref{sec:interconnection}.

\section{Locomotion Architecture}
\label{sec:method}

This section illustrates the main components of the architecture presented in Fig.~\ref{fig:architecture} and their interconnection. 

\subsection{Trajectory Generation\label{subsec:trajectory_generation}}
Given a high-level target motion to be followed, 
the objective of the trajectory generation layer is to generate the associated footstep plan, joint positions, and centroidal trajectory. 
The generation problem is framed as a nonlinear autoregressive model, 
trained on human \gls{mocap} data and fed at inference time with its own output, conditioned by the high-level target provided by an external user. 
The choice of the \gls{dnn} architecture, the data processing, and the real-time inference follow~\cite{Viceconte2022ADHERENT:Robots} and are summarized hereafter.

\subsubsection{Network architecture}
We adopt as \gls{dnn} the \gls{MANN} architecture~\cite{zhang_mode-adaptive_2018}, which is characterized by a Motion Prediction Network 
and a Gating Network. The former predicts 
the next trajectory data point given the previous one.
The latter predicts the blending coefficients which are
used at inference time to compute the weights 
of the Motion Prediction Network as a linear combination of $K$ learnable experts. 

\subsubsection{Data processing\label{sec:training_process}}
The \gls{dnn} is trained on tailored features extracted from \gls{mocap} data~\cite{Viceconte2022ADHERENT:Robots} 
retargeted onto the robot model using the Whole-Body Geometric Retargeting method~\cite{darvish_whole-body_2019} for the joint positions. 
%
%
%
%
%
To prevent the stance foot from sliding, 
the base orientation 
is directly retrieved from the human data, while the 
base position 
is computed by forward kinematics, using the lowest foot vertex as the fixed contact point.
Differently from~\cite{Viceconte2022ADHERENT:Robots}, we constrain the shoulder roll retargeting to prevent self-collisions. 
Each network input includes the motion of the base projected on the ground and retargeted joint positions and velocities. Outputs consist of subsequent base trajectory and joint configuration.

\subsubsection{Real-time inference\label{subsec:realtime_inference}}
During inference, the input from the external user is collected through a joystick. 
This user-defined base trajectory is combined with the one generated by the network. Alongside the predicted joint configuration, this merged base trajectory becomes the input for the network in the next iteration. This interactive process empowers the user to actively influence and guide the trajectory generation.

\subsection{Trajectory Adjustment}
\label{subsec:trajectory_adjustment}
The trajectory adjustment layer aims to compute feasible contact wrenches and locations while taking into account the centroidal dynamics and the output of the trajectory generation layer. Similarly to~\cite{Romualdi2022OnlineAdjustment}, the control objective is formulated here as a \gls{mpc} consisting of three fundamental components: the \emph{prediction model}, an \emph{objective function}, and a collection of \emph{inequality constraints}.
\subsubsection{Prediction Model}
\label{sec:dynamics} 
We introduce a variable $\Gamma_i \in \{0, 1\}$ to represent the contact state for a given time instant $t$. $\Gamma_i(t) = 0$ if the contact $i$ is not active at $t$, $\Gamma_i(t) = 1$ otherwise. 
Consequently, we rewrite~\eqref{eq:centroidal_dynamics} as
\begin{equation}
{}_{\bar{G}} \dot{h} =\sum_{i = 1}^{2} \sum_{j = 1}^{4} \begin{bmatrix}
      I_3 \\
      ({H}_{p_{C_i}} p_{v_{i,j}} -p_{\text{CoM}})^\wedge
    \end{bmatrix} \Gamma_i f_{i,j} + \begin{bmatrix} m g \\ 0_{3\times1} \end{bmatrix}.
        \label{eq:centroidal_dynamics_discretized}
\end{equation}
To ensure that the \gls{mpc} efficiently computes real-time control outputs, we design the optimal control problem in a way that remains agnostic to the number of active contact phases within the prediction horizon. This is accomplished by considering each contact location $p _{\mathcal{C} _ i}$ as a continuous variable evolving with the following dynamics:
\begin{equation}
    \dot{p} _{\mathcal{C} _ i}= ( 1 - \Sigma _ i)  v_{\mathcal{C} _ i},
    \label{eq:contact_dynamics}
\end{equation}
%
where $v _ {\mathcal{C}_i}$ is a slack variable,
and $\Sigma_i \in \{0, 1\}$ is related to $\Gamma_i$ as follows. When $\Gamma_i = 1$, $\Sigma_i$ must also be 1. However, $\Gamma_i = 0$ does not necessarily imply $\Sigma_i = 0$. When $\Sigma_i = 1$ then $\dot{p}_{\mathcal{C}_i} = 0$, meaning that the contact position remains constant.
The introduction of this additional parameter 
prevents contact adjustment when the contact is about to be established, differently from~\cite{Romualdi2022OnlineAdjustment}.


\subsubsection{Objective Function}
\label{sec:tasks_formulation}
The objective function consists of multiple tasks that together stabilize the dynamics~\eqref{eq:centroidal_dynamics_discretized}.
\paragraph{Centroidal trajectory tracking} To ensure that the robot follows a desired centroidal momentum trajectory, we minimize the error between the robot's angular momentum and \gls{com} trajectories with respect to the nominal as
\begin{equation}
    \Psi_{h} = \left\| \prescript{}{\bar{G}}{h}^{\omega ^n} - \prescript{}{\bar{G}}{h}^\omega \right\|^2_{\Lambda_{h}} + \left\| p_{\text{CoM}} ^ n -p_{\text{CoM}}  \right\|^2_{\Lambda_{\text{CoM}}},
    \label{eq:task_centroidal}
\end{equation}
where $\Lambda_h$ and $\Lambda_{\text{CoM}}$ are positive definite diagonal matrices.
\paragraph{Contact location regularization} To prevent the controller from computing adjusted contact locations too far from the nominal $p_{\mathcal{C}_i}^n$, we introduce the following term:
\begin{equation}
    \Psi_{p_{\mathcal{C}_i}} = \left\| p_{\mathcal{C}_i}^n  - p_{\mathcal{C}_i}  \right\|^2_{\Lambda_{p_{\mathcal{C}_i}}},
    \label{eq:task_contact}
\end{equation}
where $\Lambda_{p_{\mathcal{C}_i}}$ is a positive definite diagonal matrix.
\paragraph{Contact force regularization} To ensure that the four contact forces acting on each foot remain similar between each other and continuous, we introduce the following tasks: 
 \begin{equation}
\Psi_{f_{i,j}} = \left\| \frac{f_{i}}{n_v} - f_{i,j} \right\|^2_{\Lambda_{f_{i,j}}} + \left\| e_3^\top(f_{i}^n - f_{i}) \right\|^2_{\Lambda_{f_{i}}} +  \left\| \dot{f}_{i,j} \right\|^2_{\Lambda_{\dot{f}_{i,j}}}.
\label{eq:mpc_force}
\end{equation}
Here ${\Lambda_{\dot{f}_{i,j}}}$, $\Lambda_{f_{i,j}}$ and $\Lambda_{f_{i}}$ are positive definite diagonal matrices.
$f_{i}$ is the sum of all contact forces applied to the corners for contact $i$, while $f_{i}^n$ denotes the nominal contact force computed heuristically. The $x$ and $y$ components of the nominal force are always set to zero. The $z$ component depends on the contact's status. When the robot is in single support, this component equals the robot's weight on the contact foot and is zero on the other one. In double support, it is interpolated using a first-order spline.
In~\eqref{eq:mpc_force}, the first term penalizes the difference between the forces at different corners, the second regularizes the force to a predefined value, and the last term reduces the force rate of change.

\subsubsection{Inequality constraints} 
\label{sec:constraints}
Several inequality constraints concur to generate feasible \gls{mpc} outputs. 
\paragraph{\gls{com} height limit}
Differently from~\cite{Romualdi2022OnlineAdjustment}, we introduce a \gls{cbf} to ensure that the height of the \gls{com} remains within a safe region, $\Omega \in [p^\text{min}_{\text{CoM}_z}, p^\text{max}_{\text{CoM}_z}]$. For each time instant $t$, we define the following \gls{cbf}:
\begin{equation}
    g(t) = \texttt{-}\alpha (p^\text{min}_{\text{CoM}_z} - p_{\text{CoM}_z}) (p^\text{max}_{\text{CoM}_z} - p_{\text{CoM}_z}),
    \label{eq:cbf_com}
\end{equation}
where $\alpha > 0$, and $p_{\text{CoM}_z}$ is the z-coordinate of the \gls{com}. Eq.~\eqref{eq:cbf_com} is then used to define a set of nonlinear inequality constraints of the form~\eqref{eq:cbf_constraint}.

\paragraph{Contact Force Feasibility} To be considered feasible, the contact force must belong to the friction cone~\cite{Caron2015StabilityAreas}, which is here approximated using a conic combination of $n$ vectors. This approximation is represented as a set of linear inequalities, $A R ^\top_{\mathcal{C} _ {i} } f _ {i, j} \le b$, where $A$ and $b$ are constants determined by the static friction coefficient.
\paragraph{Contact location constraint}
To satisfy the robot's kinematic limits, we ask for the contact location to be inside a $l_b$ by $w_b$ rectangle centered on the nominal contact position and oriented as the nominal foot. This is expressed by:
\begin{equation}
l_b \le R_{\mathcal{C} _ {i} } ^\top (p_{\mathcal{C}i}^n - p_{\mathcal{C}_i}) \le u_b.
\end{equation}
\subsubsection{\gls{mpc} formulation}
\label{sec:mpc_formulation}
We tackle the \gls{mpc} problem by integrating the prediction model (Sec.~\ref{sec:dynamics}), the objective function (Sec.~\ref{sec:tasks_formulation}), and the inequality constraints (Sec.~\ref{sec:constraints}). We employ the Direct Multiple Shooting method with a constant sampling time $T_{\text{MPC}}$ \cite{BettsPractical2010}. The controller generates outputs using the Receding Horizon Principle \cite{Mayne90MPC}, with a fixed $N$-samples prediction window.

\subsubsection{\gls{mpc} and receding horizon planner \label{sec:mpc_vs_rhp}}

The formulation presented in Sec.~\ref{sec:mpc_formulation} computes the contact locations and forces.
For a torque-controlled robot, the desired contact forces can be directly set in the trajectory control layer. If the robot's available control mode is position control, 
the trajectory adjustment layer can be integrated in two ways: as a \gls{rhp} or as a proper \gls{mpc}.
In the \gls{rhp} setting, the contact forces are integrated using the centroidal dynamics~\eqref{eq:centroidal_dynamics}. The resulting \gls{com} position is provided as input to the inner layer and fed back to the planner. This method completely decouples the trajectory adjustment block from the robot, making the layer function as a planner. Conversely, in the \gls{mpc} setting, the trajectory adjustment layer receives feedback from the robot, specifically the current \gls{com} position, velocity, and angular momentum. The predicted \gls{com} quantities are then used as input to the inner layer, allowing the trajectory adjustment layer to close the loop with the robot state.

\subsection{Trajectory Control}
\label{subsec:trajectory_control}
The trajectory control layer generates the joint references to the robot and 
it consists of three main components: a CoM-ZMP controller, a swing foot planner, and a \gls{qp}-\gls{ik}.
\subsubsection{CoM-ZMP controller} The CoM-ZMP controller guarantees the tracking of the desired \gls{zmp}~\cite{vukobratovic2004zero} by using the control law~\cite{Romualdi2020ARobots}
\begin{equation}
\label{eq:ZMP_controller}
\dot{p}^*_{\text{G}} = \dot{p}^\text{ref}_{\text{G}} - K_\text{ZMP}(x_\text{ZMP}^\text{ref} - x_\text{ZMP}) + K_\text{G} (p^\text{ref}_{\text{G}} - p_{\text{G}}).
\end{equation}
Here, $p^\text{ref}_{\text{G}}$ and $p_{\text{G}}$ denote the reference and measured ground-projected \gls{com}, whereas $x^\text{ref}_{\text{ZMP}}$ and $x_{\text{ZMP}}$ refer to the reference and measured \gls{zmp}.

\subsubsection{Swing foot planner} Given a footstep list, we employ a minimum acceleration planner to generate the swing foot trajectory, as in~\cite{Romualdi2020ARobots}.

\subsubsection{QP-IK}
\label{par:qp_ik}
The joint trajectories are computed by framing the controller as a constrained optimization problem. As in~\cite{Romualdi2020ARobots}, low-priority tasks are integrated into the cost function, high-priority tasks become constraints, and the robot velocity is the optimization variable. In our formulation, the high-priority tasks include the feet poses and \gls{com} tracking.
Additionally, we introduce regularizations for the joint positions and chest orientation as low-priority tasks. We formulate the \gls{ik} as a \gls{qp} problem and solve it using off-the-shelf solvers. 
Once the joint velocity is calculated, it is integrated and sent to the robot's low-level position control.

\subsection{Joint Velocity estimation}
When the trajectory adjustment layer acts as a \gls{mpc}, reducing the noise in the estimated joints velocity is crucial to obtain smoother measurements of the \gls{com} velocity and the angular momentum. To accomplish this, we estimate the joints velocity using a \gls{kf} with the following system dynamics
\begin{equation}
    \begin{bmatrix}
        s_{k+1} \\
        \dot{s}_{k+1} \\
        \ddot{s}_{k+1}
    \end{bmatrix} = 
      \begin{bmatrix}
        1 & T & 0 \\
        0 & 1 & T \\
        0 & 0 & 1
    \end{bmatrix}
        \begin{bmatrix}
        s_{k} \\
        \dot{s}_{k} \\
        \ddot{s}_{k}
    \end{bmatrix}
    + v.
\end{equation}
$T$ is the sampling time, and $v\sim\mathcal{N}(0,Q)$ 
is the process noise. 
The measurement equation is $ y_k = s_{k}  + w$, where $w\sim\mathcal{N}(0,R)$. The covariance state at $t=0$ is $P_0 = \lambda I_3$.
\par
To reduce the need for manual hand-tuning, we identify the \gls{kf} covariance matrices by maximizing an objective function:
\begin{equation}\label{eq:kf_tuning}
    \xi ^* = \argmax\limits_{\xi} \mathcal{G}(\xi).
\end{equation}
$\xi \in \mathbb{R}^5_+$ is the vector containing the initial state covariance parameter $\lambda\in\mathbb{R}_+$, the diagonal terms of the covariance matrix $Q\in\mathbb{R}^{3\times 3}$, and the value of the measurement covariance $R\in\mathbb{R}_+$. 
%
To compute $\xi$, we collect a joint position dataset and, for each point, compute the estimated output of the \gls{kf} along with the following objective function $\mathcal{G}_i(\xi)$:
\begin{equation}
    \label{eg:kf_ga_of}
    \mathcal{G}_i(\xi) = \texttt{-}(w_{j} \dddot{s}^2_i + w_{a} \ddot{s}^2_i + (s_i - \hat{s}_i)^2 + (s_i - \hat{s}_i^v)^2 + (s_i - \hat{s}_i^a)^2).
\end{equation}
Here, $w_{a}$ and $w_{j}$ are positive weights, and $\hat{(\cdot)}$ indicates the quantities estimated by the \gls{kf}. $\hat{s}_i^v$ and $\hat{s}_i^a$ are the joint positions numerically integrated from the estimated velocity and acceleration, respectively.  
In~\eqref{eg:kf_ga_of}, the first two terms encourage the continuity of the estimated quantities by minimizing the jerk and the acceleration, while the last three terms drive the \gls{kf} to estimate velocity and acceleration such that their integration does not drift from the position.
The objective function $\mathcal{G}(\xi)$ in \eqref{eq:kf_tuning} is computed as the sum of the objective functions for each timestep and optimized using \gls{ga}.\looseness=-1
\subsection{System integration}
\label{sec:interconnection}

This section addresses the challenges introduced in Sec.~\ref{sec:problem_statement} and further describes the interconnections between the layers of our architecture in Fig.~\ref{fig:architecture}.
\subsubsection{\hyperref[c1]{$\mathscr{C}_1$}}
The main objective of the trajectory generation layer is to compute the \emph{nominal} \gls{com}, angular momentum, and footsteps used in the \gls{mpc} objective function — Eqs.~\eqref{eq:task_centroidal}, \eqref{eq:task_contact}, \eqref{eq:mpc_force}. However, the autoregressive \gls{dnn} described in Sec.~\ref{subsec:trajectory_generation} predicts only the joint states and the base pose. So, we compute the centroidal quantities by applying the forward kinematics. The contact states are inferred through a Schmitt trigger. 
When a corner of the foot is below a given threshold for a predefined amount of time, we consider the contact active. We then determine the contact location through forward kinematics, zeroing the height of the foot.
%
As the length of the generated footsteps may require torques that closely approach the joint torque limits, we iteratively scale the generated footsteps and \gls{com} positions as follows:
\begin{equation}
p^s_{\circ_i} = p^s_{\circ_{i-1}} + \gamma (p^n_{\circ_{i}} - p^n_{\circ_{i-1}}).
\label{eq:scaling}
\end{equation}
Here, $p_\circ$ represents either the contact or the \gls{com} position. The indices $i$ and $i-1$ correspond to two consecutive time steps for the \gls{com} or two consecutive footsteps in the case of contacts. The superscripts $s$ and $n$ denote the scaled and nominal quantities, respectively. The parameter $\gamma > 0$ is the scaling factor.
Similarly, the centroidal trajectory and contact timings are time-scaled by a positive factor $\eta$. The higher $\eta$, the slower the trajectory.
\subsubsection{\hyperref[c2]{$\mathscr{C}_2$}}
The \gls{dnn} computes the output for the next time step, but the trajectory adjustment layer requires a trajectory over an entire horizon. To bridge this gap, the \gls{dnn} inference is executed iteratively, generating a trajectory that matches the duration specified by the \gls{mpc}'s horizon while keeping the joypad input constant -- Sec.~\ref{subsec:realtime_inference}. 
Since the \gls{dnn} might be trained with a dataset collected at a different frequency than the \gls{mpc}, 
we need to synchronize the calls of the trajectory generation and adjustment layers.
To address this, we generate trajectories for the \gls{mpc} horizon
in a separate process 
at a rate $\delta_\text{DNN}$ determined by the least common multiple of the \gls{mpc} sampling time, $T_\text{MPC}$, and the scaled \gls{dnn} period, $\eta T_\text{DNN}$.
This approach minimizes the frequency of trajectory generator calls, improving the system's overall performance. 
The network outputs are stored for the entire horizon, and the sample at index $\delta_\text{DNN} / \eta T_\text{DNN}$ is used to reset the \gls{dnn} when a new trajectory is required.
\subsubsection{\hyperref[c3]{$\mathscr{C}_3$}}
To provide to the \gls{mpc} references at its expected rate, we resample the generated trajectories at the \gls{mpc} sampling time using a first-order spline.
%
%
\subsubsection{Interconnections}
As regards the interconnection between the trajectory generation and adjustment layer, 
we initialize the latter with nominal values of \gls{com}, angular momentum, and contact locations provided by the former. Depending on whether the trajectory adjustment layer functions as a \gls{mpc} or a \gls{rhp}, its feedback varies. For the \gls{mpc}, the feedback is the robot state. For the \gls{rhp}, the feedback consists of the previously-integrated desired centroidal quantities. To make the \gls{rhp} aware of external disturbances affecting the robot, the measured external wrench is used as a disturbance in the prediction model -- Eq.~\eqref{eq:centroidal_dynamics_discretized} 
\par
Regarding the interconnection between the trajectory adjustment and control layers, we feed the adjusted footstep list produced by the former to the swing foot planner. The planner can update the foot trajectory with new desired contact positions even during the swinging phase. Moreover, the desired \gls{com} for the \gls{com}-\gls{zmp} controller~\eqref{eq:ZMP_controller} is obtained from the \gls{mpc}'s predicted state when the trajectory adjustment layer acts as a controller, or integrated from computed contact forces when it acts as \gls{rhp}. The desired \gls{zmp} is computed from the desired contact forces provided by the trajectory adjustment layer.
%
Finally, the trajectory generation layer is directly connected to the trajectory control layer to provide the nominal joint trajectories which serve as a regularizer in the \gls{qp}-\gls{ik} problem.

\section{Results}
\label{sec:results}

In this section, we present the validation results for the architecture discussed in Sec.~\ref{sec:method}.
Our implementation of \gls{MANN} features a Motion Prediction Network and a Gating Network, each with 3 hidden layers of 512 and 32 units, respectively, using $K=4$ experts. We consider $n=26$ joints. Human \gls{mocap} data is collected using an Xsens suit~\cite{roetenberg_xsens_2013}, containing 17 inertial sensors. The data includes 20 minutes of walking motions and transitions on flat terrain at $T_\text{DNN} = \SI{20}{\milli\second}$, resulting in 150k training points when mirrored. The network, implemented in PyTorch~\cite{paszke_pytorch_2019} and trained using AdamWR~\cite{loshchilov_fixing_2017} for 110 epochs, takes about 7 hours on an NVIDIA GeForce GTX 1650 GPU for training. Online inference uses ONNX Runtime~\cite{onnx_runtime_developers_onnx_2021}.
For the non-linear \gls{mpc} optimization problem, we use CasADi 3.6.2~\cite{Andersson2018CasADi:Control} and IPOPT 3.14.12~\cite{IPOpt2006} with HSL\_MA97~\cite{Hogg2011HSL_MA97Systems} libraries. The \gls{ik}-\gls{qp} problem in the trajectory control layer is solved with osqp-eigen and the OSQP library~\cite{Stellato2018}.
Kalman Filter gains are optimized using a \gls{ga} with PyGAD~\cite{Gad2023PyGAD:Library}, running for 50 generations with 60 parents selected for mating. The population size is 120, using $k$-tournament selection ($k=4$) and a two-point crossover strategy. A 20\% per gene mutation rate with random mutations is used, preserving the top 10\% of solutions in each generation. Data for the \gls{ga} consist of a 6-minute trajectory sampled at \SI{1}{\kilo\hertz}, for a total of about 360k points. Optimizing the values for all 26 controlled joints takes around 7 hours, averaging 30 minutes per joint, on an AMD EPYC 7513 32-core Processor.
The architecture is implemented in C++ and the code is available at~\url{https://sites.google.com/view/dnn-mpc-walking}.
\par
To validate the performance of the proposed architecture, we conduct two main experiments involving forward walking and turning. In the first experiment, the trajectory adjustment layer operates as a \gls{rhp}, while in the second one, it functions as a \gls{mpc}. Both experiments, shown in the accompanying video, are conducted on the \SI{160}{\centi\meter} tall, \SI{56}{\kilo\gram} ergoCub humanoid robot. The architecture runs on a 10th-generation Intel Core i7-10750H laptop running Ubuntu Linux 22.04.

\subsection{Trajectory Adjustment layer as \gls{rhp}}
\label{subsec:res_rhp}

\begin{figure}
    \centering
    \includegraphics[width=\columnwidth]{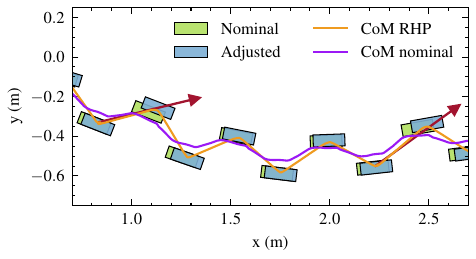} 
    \caption{Top-view of the walking pattern performed by ergoCub with the trajectory adjustment layer acting as \gls{rhp}. The robot is pushed twice. The red arrows represent the external pushes.}
    \label{fig:steps_and_com_p}
    \vspace{-0.5cm}
\end{figure}

In this experiment, the trajectory adjustment layer operates with a sampling time $T_\text{MPC} = \SI{60}{\milli\second}$ and $\SI{1.2}{\second}$ horizon. The trajectory control runs at $\SI{500}{\hertz}$. The nominal footsteps are scaled by $\gamma=0.7$ and $\eta=3$, resulting in $\delta_\text{DNN} = \SI{60}{\milli\second}$.
\par
Fig.~\ref{fig:steps_and_com_p} shows the nominal and adjusted footsteps, the nominal \gls{com}, and the \gls{com} computed from the integration of the \gls{rhp} output. The nominal \gls{com}, used as a warm start for the \gls{rhp}, undergoes significant modifications to meet force feasibility constraints and the objective function in Sec.~\ref{subsec:trajectory_adjustment}. 
Moreover, since the step adjustment is always considered in the \gls{rhp} formulation, the nominal footsteps are continuously modified. A more pronounced step adjustment emerges when the robot is perturbed with an external impulsive force applied to its right arm during walking, as indicated by the red arrows in Fig.~\ref{fig:steps_and_com_p}.
Being the trajectory adjustment layer implemented as a \gls{rhp}, it is not aware of \gls{com} perturbances.
%
To address this, we incorporate the estimated external force as a measured disturbance in the centroidal dynamics in Eq.~\eqref{eq:centroidal_dynamics_discretized} as well as in the contact forces integrator of Sec.~\ref{sec:mpc_vs_rhp}.
Given the impulsive nature of the external force, we consider the disturbance to be non-zero only for the first sample of the \gls{rhp} horizon, setting it to zero for the remaining window.
The external force is estimated online considering the readouts of the Force Torque sensors mounted on the robot arms and the robot state~\cite{Nori2015}. 
The \gls{rhp} compensates for the disturbance effect by adjusting the footstep locations up to \SI{5}{\centi\meter} under a \SI{63}{\newton} impulsive force.

\begin{figure}
    \centering
    \includegraphics[width=\columnwidth]{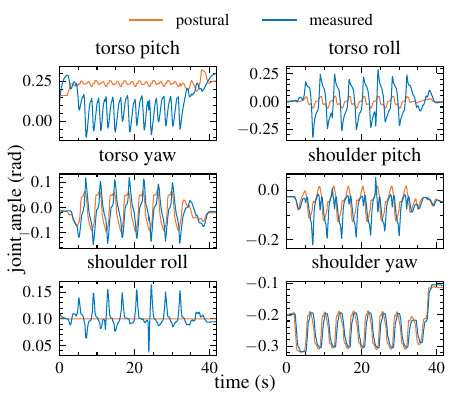} 
    \caption{Measured upper body joint angles - specifically, torso and left arm - compared with their associated data-driven postural during walking.}
    \label{fig:joints_tracking}
    \vspace{-0.5cm}
\end{figure}

In Fig.~\ref{fig:joints_tracking}, we examine how closely the measured joint positions follow their associated data-driven posturals. 
The upper body joints display periodic patterns similar to human data, unlike traditional humanoid locomotion architectures that use a fixed postural reference~\cite{Viceconte2022ADHERENT:Robots}. 
As the \gls{qp}-\gls{ik} incorporates a task to zero the chest roll and pitch angles, the measured values for the torso joints exhibit larger deviations from their associated postural compared to the other joints. The shoulder roll postural remains constant because of the retargeting constraint to prevent self-collisions.  
Overall, the upper body motion contributes to shaping the robot's movement to resemble human \gls{mocap} data, as shown in the accompanying video.

\subsection{Trajectory Adjustment layer as \gls{mpc}}
\label{subsec:res_mpc}
When the trajectory adjustment layer implements a closed-loop \gls{mpc}, it is necessary to reduce the \gls{mpc} sampling time to $T_\text{MPC} = \SI{50}{\milli\second}$ to achieve smoother and faster reactions to external disturbances, while the trajectory control operates at \SI{500}{\hertz}. As before, the MPC time horizon is $\SI{1.2}{\second}$. In this experiment, the nominal footsteps are scaled 
by factors $\gamma=0.6$ and $\eta=3$, respectively, therefore $\delta_\text{DNN} = \SI{300}{\milli\second}$.
\par
\begin{figure}
    \centering
    \includegraphics[width=\columnwidth]{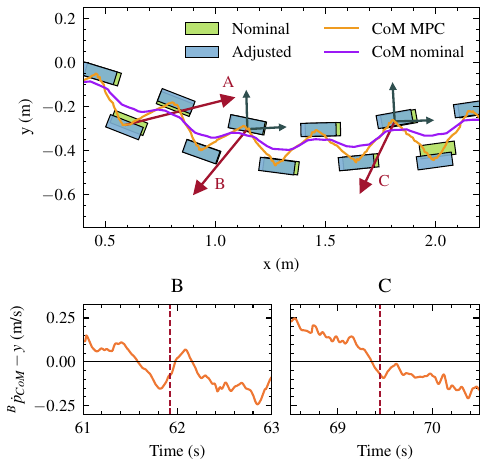} 
    \caption{Top-view of the walking pattern performed by ergoCub with the trajectory adjustment layer functioning as a model predictive controller \gls{mpc}. The robot is pushed three times, as indicated by the red arrows. For the second (B) and third (C) disturbances, occurring at \SI{61.9}{\second} and \SI{69.43}{\second} respectively, the lateral (local y-coordinate) \gls{com} velocity is displayed. The red dashed line marks the time at which the system is perturbed.}
    \label{fig:steps_and_com_c}
    \vspace{-0.5cm}
\end{figure}

The top plot of Fig.~\ref{fig:steps_and_com_c} represents the nominal and adjusted footsteps, the nominal and predicted \gls{com}. As in the previous experiment, the nominal \gls{com}, which serves as a warm start for the \gls{mpc}, undergoes significant modifications to meet the force feasibility constraints and optimize the objective function. With the trajectory adjustment layer implemented as a \gls{mpc}, it now adjusts the contact foot positions based on the \gls{com} state.
Three external disturbances, labeled A, B, and C, affect the system. Each disturbance impacts the robot's arm, which behaves compliantly. During disturbance A, the robot experiences a push on its shoulder. Since only the shoulder is compliant and reaches its joint limit, the change in the \gls{com} is limited, resulting in a small footstep adjustment. In contrast, disturbances B and C involve pulling forces on the robot. Although the pulling forces in B and C are comparable (about \SI{68}{\newton}), step recovery is triggered only after C. This difference arises because the \gls{mpc} is integrated within a three-layer architecture, where also the trajectory control layer can adjust the \gls{com} trajectory at a higher frequency to prevent the robot from falling. In scenario B, the step adjustment is not triggered because the body \gls{com} velocity ${}^B \dot{p}_\text{CoM}$ on the lateral plane at $t \approx \SI{61.90}{\second}$ is positive, i.e. opposite to the external disturbance on the y coordinate, indicating that the robot is not falling. Conversely, in scenario C, the body \gls{com} velocity ${}^B \dot{p}_\text{CoM}$ on the lateral plane at $t \approx \SI{69.43}{\second}$ is negative, i.e. aligned with the external disturbance, suggesting that the robot is about to fall inward. Therefore, the step recovery is automatically triggered.
\par
\begin{figure}
    \centering
    \includegraphics[width=\columnwidth]{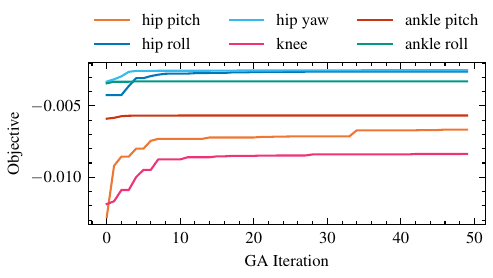} 
    \caption{Values of the objective function across \gls{ga} iterations for the \gls{kf} parameters estimation of the left leg.}
    \label{fig:kf_ga}
    \vspace{-0.5cm}
\end{figure}

In Fig.~\ref{fig:kf_ga}, we present the values of the objective function~\eqref{eg:kf_ga_of} for several joints of the left leg across different \gls{ga} iterations. The plot demonstrates that, after approximately 20 iterations, the objective function values converge to their maximum, indicating successful optimization. Similar convergence behavior is observed for the other joints as well.


\subsection{\gls{rhp} vs \gls{mpc}}


The experiments in Sec.~\ref{subsec:res_rhp} and Sec.~\ref{subsec:res_mpc} reveal different behaviors when the trajectory adjustment layer functions as a \gls{rhp} or as a \gls{mpc}.
In the case of \gls{rhp}, since the feedback is derived from desired quantities, the layer is unaffected by measurement noise, allowing for longer steps. However, to prevent falls from external pushes, \gls{rhp} must explicitly consider external disturbances, requiring reliable force measurements. Conversely, with \gls{mpc}, no force measurements are needed as it uses the robot's kinematic state as feedback. This feedback can be noisy due to joint velocity estimation errors, necessitating extensive tuning. Our \gls{ga}-tuned \gls{kf} mitigates this issue, making the trajectory adjustment acting as \gls{mpc} the best compromise for a complete integrated locomotion architecture.\looseness=-1
\section{Conclusion}
\label{sec:conclusions}



This paper contributes towards bridging the gap between model-based trajectory adjustment and data-driven trajectory generation for humanoid robot locomotion. The trajectory adjustment layer implements a \gls{mpc} or a \gls{rhp} and ensures dynamically-feasible \gls{com} motion while addressing the step adjustment problem. The two implementations are benchmarked. In the case of \gls{mpc}, we introduced a \gls{ga}-tuned \gls{kf} to reduce joint velocity noise and consequently \gls{com} velocity and angular momentum noise. Results on the ergoCub humanoid robot indicate that the architecture prevents falls, replicates human \gls{mocap} walking styles, and withstands disturbances up to 68 N.
\par
The current system lacks a base estimator, limiting its ability to handle stronger pushes. Adding a base estimator would improve fall detection and consequently contact position adjustment. Transitioning to torque control would also enhance natural responses to disturbances. Future work will focus on integrating a base estimator and torque control to improve robustness and enable handling more significant perturbations.\looseness=-1


\bibliography{references, zotero_references, MyBSTcontrol}

\bibliographystyle{IEEEtran}

\end{document}